\documentclass{article}

\usepackage{microtype}
\usepackage{graphicx}
\usepackage{subfigure}
\usepackage{booktabs} %

\usepackage{hyperref}

\usepackage{microtype}
\usepackage{graphicx}
\usepackage{subfigure}
\usepackage{booktabs} %
\usepackage{xcolor}

\usepackage{hyperref}

\usepackage{pgfplots}
\pgfplotsset{compat = newest}
\usepackage{multirow}

\usepackage{amsmath}
\usepackage{amssymb}
\usepackage{mathrsfs}
\usepackage{mathtools}
\usepackage{amsthm}
\usepackage{algorithmic}
\usepackage{algorithm}
\usepackage{listings}
\usepackage{makecell}
\usepackage{colortbl}
\usepackage{color}
\usepackage{wrapfig}
\usepackage{cancel}
\usepackage{soul,xcolor}
\usepackage{pifont}
\usepackage{tikz}
\usetikzlibrary{shapes, positioning, arrows.meta, calc, decorations.pathmorphing, quotes}

\usepackage[capitalize,noabbrev]{cleveref}

\theoremstyle{plain}
\newtheorem{theorem}{Theorem}[section]

\newtheorem{lemma}[theorem]{Lemma}

\theoremstyle{definition}
\newtheorem{definition}[theorem]{Definition}

\theoremstyle{remark}

\usepackage[textsize=tiny]{todonotes}

\newcommand{\alink}[1]{\href{#1}{paper-link}}

\usepackage{enumitem}
\usepackage{ amssymb }

\definecolor{codebg}{rgb}{0.95,0.95,0.95}
\definecolor{codeblue}{rgb}{0.13,0.13,1}
\definecolor{codegreen}{rgb}{0,0.5,0}
\definecolor{codegray}{rgb}{0.5,0.5,0.5}
\definecolor{codered}{rgb}{0.7,0,0}

\definecolor{citecolor}{HTML}{0071BC}
\definecolor{linkcolor}{HTML}{ED1C24}
\hypersetup{colorlinks=true, linkcolor=linkcolor, citecolor=citecolor,urlcolor=black}

\usepackage{amsmath,amsfonts,bm}

\def\eqref#1{equation~\ref{#1}}

\def\1{\bm{1}}

\DeclareMathAlphabet{\mathsfit}{\encodingdefault}{\sfdefault}{m}{sl}
\SetMathAlphabet{\mathsfit}{bold}{\encodingdefault}{\sfdefault}{bx}{n}

\DeclareMathOperator*{\argmax}{arg\,max}
\DeclareMathOperator*{\argmin}{arg\,min}

\renewcommand{\cite}{\citep}
\usepackage[accepted]{icml2024}

\usepackage{amsmath}
\usepackage{amssymb}
\usepackage{mathtools}
\usepackage{amsthm}

\usepackage[capitalize,noabbrev]{cleveref}

\usepackage[textsize=tiny]{todonotes}

\icmltitlerunning{OTMatch: Improving Semi-Supervised Learning with Optimal Transport}

\begin{document}

\twocolumn[
\icmltitle{OTMatch: Improving Semi-Supervised Learning with Optimal Transport}

\icmlsetsymbol{equal}{*}

\begin{icmlauthorlist}
\icmlauthor{Zhiquan Tan}{thu}
\icmlauthor{Kaipeng Zheng}{sjtu}
\icmlauthor{Weiran Huang \textsuperscript{\dag}}{sjtu,ailab}
\end{icmlauthorlist}

\icmlaffiliation{sjtu}{MIFA Lab, Qing Yuan Research Institute, SEIEE, Shanghai Jiao Tong University}
\icmlaffiliation{thu}{Department of Mathematical Sciences, Tsinghua University}
\icmlaffiliation{ailab}{Shanghai AI Laboratory}

\icmlcorrespondingauthor{Weiran Huang}{weiran.huang@outlook.com}

\icmlkeywords{Machine Learning, ICML}

\vskip 0.3in
]

\printAffiliationsAndNotice{}  %

\begin{abstract}
Semi-supervised learning has made remarkable strides by effectively utilizing a limited amount of labeled data while capitalizing on the abundant information present in unlabeled data. 
However, current algorithms often prioritize aligning image predictions with specific classes generated through self-training techniques, thereby neglecting the inherent relationships that exist within these classes. 
In this paper, we present a new approach called OTMatch, which leverages semantic relationships among classes by employing an optimal transport loss function to match distributions. We conduct experiments on many standard vision and language datasets. The empirical results show improvements in our method above baseline, this demonstrates the effectiveness and superiority of our approach in harnessing semantic relationships to enhance learning performance in a semi-supervised setting.
\end{abstract}

\section{Introduction}

Semi-supervised learning occupies a unique position at the intersection of supervised learning and self-supervised learning paradigms ~\citep{tian2020contrastive, chen2020simple}. 
The fundamental principle behind semi-supervised learning lies in its ability to leverage the latent patterns and structures of a substantial amount of unlabeled samples to collaborate with conventional supervised learning on labeled samples.
It has demonstrated remarkable performance without the need for extensive human efforts in data collection~\citep{sohn2020fixmatch,zhang2021flexmatch,wang2022freematch}. 

Pseudo-labeling-based methods have dominated the research in semi-supervised learning. 
It dynamically assigns labels to unlabeled samples to prepare an extended dataset with labeled samples for model training~\citep{lee2013pseudo,tschannen2019mutual,berthelot2019mixmatch,xie2020unsupervised,sohn2020fixmatch,gong2021alphamatch,zhang2021flexmatch,wang2022freematch}, allowing the model to benefit from the potential knowledge contained in these unlabeled samples.
Typically, pseudo-labels are derived from the neural network's confidence, and the cross-entropy loss is used to align the prediction on the strongly augmented image view with the generated pseudo-labels.
Following this paradigm, recent works have achieved state-of-the-art performance~\citep{zhang2021flexmatch,wang2022freematch} in semi-supervised learning.
However, it has been demonstrated that the overconfidence of models can be observed when assigning pseudo labels during training~\citep{overconfidence}.
This means that the model could confidently assign samples to incorrect categories, due to the limited labeled samples in semi-supervised learning.
Such unreliability in confidence is particularly pronounced when dealing with extremely limited labeled samples.
As a result, those misclassified samples can misguide the model's learning process, resulting in a decrease in performance.

In this work, we propose a novel solution to alleviate the issue of overconfidence in the model in pseudo-labeling-based methods by incorporating inter-class semantic relationships.
Particularly, we find that a key issue lies in simply employing the traditional cross-entropy loss to train the model with the pseudo-labeled samples.
In the cross-entropy loss, the representation of a sample is aligned with a single category. 
However, when a sample is assigned an incorrect pseudo-label, this leads the model to learn the representation in the wrong direction.
To address this issue, for pseudo-labeled samples, we propose to incorporate comprehensive inter-class relationships for regularization, instead of a single-category target, to guide model training, hence offering improved robustness.
Particularly, we use optimal transport to tackle this problem as the predicted probability and pseudo labels are naturally two distributions that need to be matched, and semantic information can be injected into the cost matrix in optimal transport loss.

\begin{figure*} 
\centering 
\includegraphics[width=14cm, height=6cm, trim=0cm 4cm 0cm 0cm, clip]{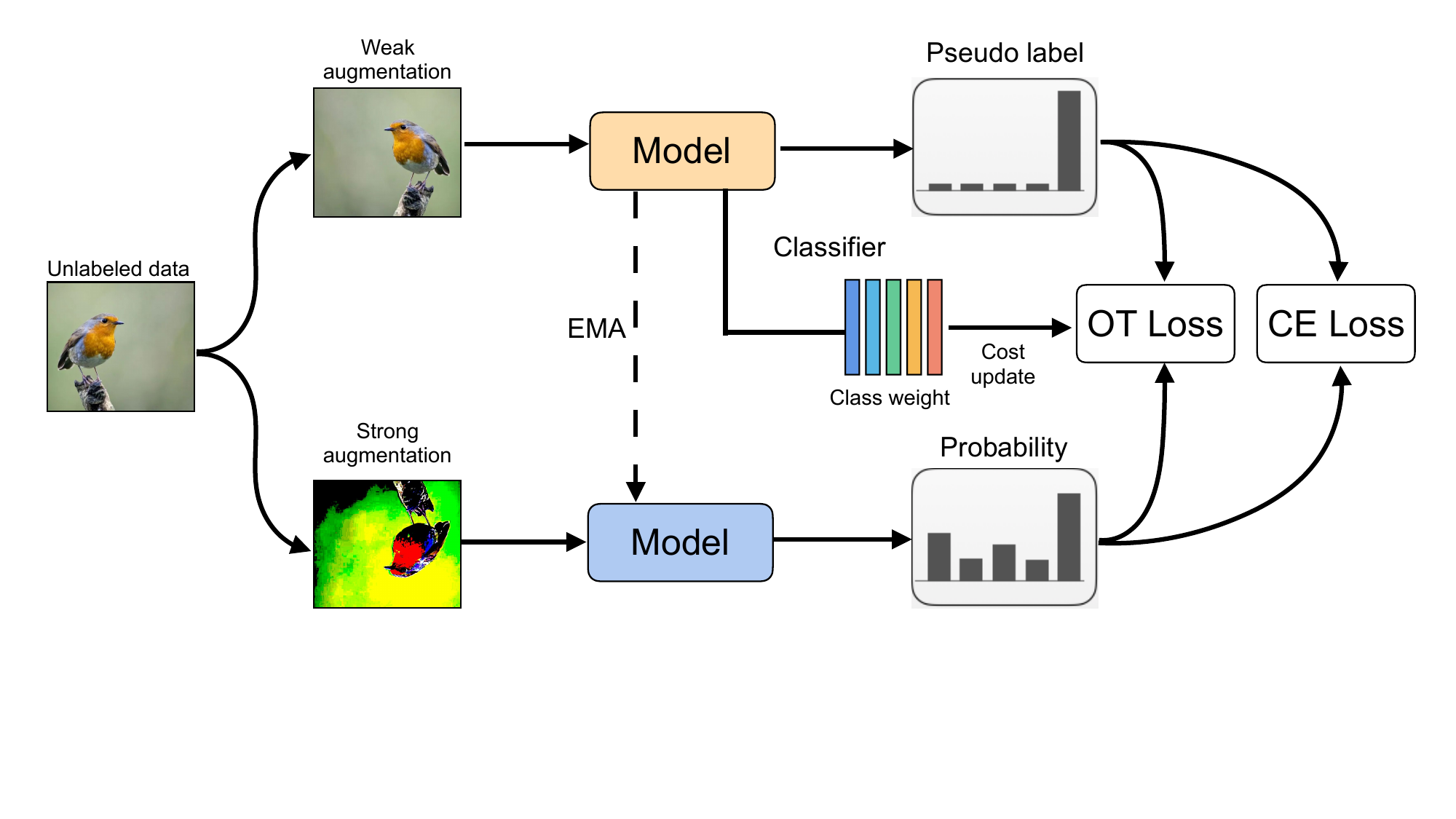}
\caption{To obtain a pseudo-label, a model is fed with a weakly augmented image. Then, the model predicts the probability of a strongly augmented version of the same image. The loss includes cross-entropy and optimal transport loss, which considers the probability and pseudo-label. The cost used in optimal transport is adjusted based on the model's classification head weight.}
\label{fig: arch}
\end{figure*}

The paper is organized in the following manner: We present a new perspective on current semi-supervised learning methods that rely on pseudo-labeling. We view these methods as aiming to align the semantic distribution captured by the teacher and student models, motivated by \citep{shi2023understanding}. 
We further improve this by proposing a novel training algorithm OTMatch, where we bootstrap the cost in optimal transport from the class embedding throughout model training, thus fully capitalizing on semantic relationships between classes.
The diagram of OTMatch is presented in Figure \ref{fig: arch}. 

Our contributions can be summarized as follows:
\vspace{-0.2cm}
\begin{itemize}%

\item We provide a novel understanding of current pseudo-labeling-based semi-supervised learning methods by viewing them as matching the distribution of semantics obtained by the teacher and student models using (inverse) optimal transport. We also extend this framework to analyze algorithms in self-supervised learning. 

\item We propose OTMatch, a novel semi-supervised learning algorithm that exploits the semantic relationship between classes to alleviate the issue of model overconfidence caused by limited labeled samples.

\item We carry out experiments on well-known vision datasets such as CIFAR 10/100, STL-10, and ImageNet, noting that our method shows improvements, particularly in challenging situations with very few labeled samples. Additionally, we perform experiments in the language modality, discovering that our method is effective there as well.
\end{itemize}

\section{Related Work}

Semi-supervised learning, aiming to enhance model performance through the utilization of abundant unlabeled data, has attracted considerable attention in recent years~\citep{chen2020big, assran2021semi, wang2021data, zhang2023relationmatch, chen2023boosting, nassar2023protocon, huang2023flatmatch}. 
The invariance principle serves as the foundation for many effective semi-supervised algorithms. 
Essentially, this principle posits that two semantically similar images should yield similar representations when processed by a same backbone.

\textbf{Consistency regularization.} 
Consistency regularization, initially introduced in the $\Pi$-Model~\citep{Rasmus2015SemiSupervisedLW}, has emerged as a prevalent technique for implementing the invariance principle. 
This method has gained widespread adoption in subsequent research~\citep{tarvainen2017mean, laine2016temporal, berthelot2019mixmatch}. Consistency regularization entails the generation of pseudo-labels and the application of appropriate data augmentation strategies~\citep{tschannen2019mutual, berthelot2019mixmatch, xie2020unsupervised, sohn2020fixmatch, gong2021alphamatch}. Pseudo-labels are created for unlabeled data and utilized in subsequent training iterations~\citep{lee2013pseudo}. 
The conventional approach involves minimizing the cross-entropy objective to align the predicted pseudo-labels of two distorted images, typically obtained through data augmentation ~\citep{Rasmus2015SemiSupervisedLW, laine2016temporal, tarvainen2017mean}. 
Extensive research has recently focused on generating efficient and informative pseudo-labels \citep{hu2021simple, nassar2021all, xu2021dash, zhang2021flexmatch, li2022maxmatch, wang2022debiased}, achieving state-of-the-art performance.
SimMatch \citep{zheng2022simmatch} and CoMatch \citep{li2021comatch} also investigate contrastive learning for consistency regularization.
The efficacy of consistency regularization has been demonstrated as a simple yet effective approach, serving as a foundational component in numerous state-of-the-art semi-supervised learning algorithms. 

\textbf{Improving pseudo-label quality.}
In the realm of semi-supervised learning, the current discourse surrounding consistency regularization primarily revolves around augmenting the quality of pseudo-labels.
SimPLE~\citep{hu2021simple} introduces a paired loss function that diminishes the statistical discrepancy between confident and analogous pseudo-labels, thereby enhancing their quality. 
Dash~\citep{xu2021dash} and FlexMatch~\citep{zhang2021flexmatch} propose dynamic and adaptable filtering techniques for pseudo-labels, which are better suited for the training process. 
CoMatch~\citep{li2021comatch} advocates for the integration of contrastive learning within the framework of semi-supervised learning, enabling the simultaneous learning of two representations of the training data. 
SemCo~\citep{nassar2021all} takes into account external label semantics to safeguard against pseudo-label quality deterioration for visually similar classes, employing a co-training approach. 
FreeMatch~\citep{wang2022freematch} proposes a self-adjusting confidence threshold that considers the learning status of the models, allowing for improved control over pseudo-label quality. 
MaxMatch~\citep{li2022maxmatch} presents a consistency regularization technique that minimizes the most substantial inconsistency between an original unlabeled sample and its multiple augmented versions, accompanied by theoretical guarantees. 
NP-Match~\citep{wang2022np} employs neural processes to amplify the quality of pseudo-labels. 
SEAL~\citep{tan2023seal} introduces a methodology that facilitates the concurrent learning of a data-driven label hierarchy and the execution of semi-supervised learning. 
SoftMatch~\citep{chen2023softmatch} addresses the inherent trade-off between the quantity and quality of pseudo-labels by utilizing a truncated Gaussian function to assign weights to samples based on their confidence. 

Unlike previous works focusing on enhancing pseudo-label quality,
we address the issue of overconfidence in models from an orthogonal perspective by incorporating inter-class relationships as constraints.
\citet{taherkhani2020transporting, tai2021sinkhorn} also explore using optimal transport in semi-supervised learning.
However, they still use optimal transport for pseudo-label filtering.
In contrast, we employ optimal transport theory to provide a novel understanding of current pseudo-labeling methods. 
We further improve this by extracting inter-class semantic relationships from the model's learning dynamic to update the cost matrix, which has never been explored before.

\section{Preliminary}

\subsection{Problem setting and notations}

Throughout a semi-supervised learning process, it is customary to have access to both labeled and unlabeled data. Each batch is a mixture of labeled data and unlabeled data. Assume there are $B$ labeled samples $\{ (\mathbf{x}_{l_i}, \mathbf{y}_{l_i}) \}^{B}_{i=1}$ and $\mu B$ unlabeled samples $\{ \mathbf{x}_{u_i} \}^{\mu B}_{i=1}$ in a mixed batch, where $\mu$ is the ratio of unlabeled samples to labeled samples. We adopt the convention in semi-supervised learning \citep{zhang2021flexmatch,wang2022freematch} that there will be a teacher and student network that shares the same architecture. The teacher network does not update through gradient but by exponential moving average (EMA) instead. There will also be two sets of augmentations of different strengths, namely weak augmentation $\omega(\cdot)$ and strong augmentation $\Omega(\cdot)$. 

For labeled data, the loss is the classical cross-entropy loss as follows:
\begin{equation}
\label{L_sup}
\mathcal{L}_{\text{sup}} = \frac{1}{B} \sum^B_{i=1} \text{H}(\mathbf{y}_{l_i}, \text{Pr}(\omega(\mathbf{x}_{l_i}))),   
\end{equation}
where $\text{Pr}(\omega(\mathbf{x}_{l_i}))$ denotes the output probability and $\text{H}(\cdot, \cdot)$ denotes the cross-entropy loss.

The unsupervised loss $L_{un}$ is usually the main focus of improving semi-supervised learning. FixMatch \citep{sohn2020fixmatch} introduces the idea of using a fixed threshold $\tau$ to only assign pseudo labels to those samples with enough confidence. Later, a line of works like FlexMatch \citep{zhang2021flexmatch} and FreeMatch \citep{wang2022freematch} seeks to improve the threshold selection strategy. This loss can be formally described as follows: 
\begin{equation}
\label{L_un1}
\mathcal{L}_{\text{un1}} = \frac{1}{\mu B} \sum^{\mu B}_{i=1} \mathbf{I}( \text{max}(\mathbf{q}_{u_i})> \tau) \text{H}(\hat{\mathbf{q}}_{u_i}, \mathbf{Q}_{u_i}),    
\end{equation}
where $\mathbf{q}_{u_i}$ is the probability of (the teacher) model on the weakly-augmented image, $\mathbf{Q}_{u_i}$ is the probability of (the student) model on the strongly-augmented image, and $\hat{\mathbf{q}}_{u_i}$ denotes the generated one-hot hard pseudo label.

FreeMatch \citep{wang2022freematch} also introduces a fairness loss to make the class distribution more balanced.
The loss is given as follows:
\begin{equation}
\label{L_un2}
\mathcal{L}_{\text{un2}} =  - \text{H}(\text{SumNorm}(\frac{\mathbf{p}_1}{\mathbf{h}_1}), \text{SumNorm}(\frac{\mathbf{p}_2}{\mathbf{h}_2})), \end{equation}
where $\mathbf{p}_1$ and $\mathbf{h}_1$ denote the mean of model predictions and histogram
distribution on weakly-augmented images respectively, $\mathbf{p}_2$ and $\mathbf{h}_2$ is defined on the pseudo-labeled strongly-augmented images. As the prediction on the weakly-augmented image is more accurate, using cross-entropy loss here mimics the maximization of entropy.

\subsection{Optimal transport}

The Kantorovich formulation of discrete optimal transport \cite{kantorovich1942transfer}, also known as the transportation problem, provides a mathematical definition for finding the optimal transportation plan between two discrete probability distributions. Let's consider two discrete probability distributions, denoted as $\mu$ and $\nu$, defined on two finite sets of points, $\mathbf{X} = \{\mathbf{x}_1, \mathbf{x}_2, \ldots, \mathbf{x}_m\}$ and $\mathbf{Y} = \{\mathbf{y}_1, \mathbf{y}_2, \ldots, \mathbf{y}_n\}$, respectively. The goal is to find a transportation plan that minimizes the total transportation cost while satisfying certain constraints. The transportation plan specifies how much mass is transported from each point in $\mathbf{X}$ to each point in $\mathbf{Y}$. This is achieved by defining a transportation (plan) matrix $\mathbf{T} = [\mathbf{T}_{ij}]$, where $\mathbf{T}_{ij}$ represents the amount of mass transported from point $\mathbf{x}_i$ to point $\mathbf{y}_j$.

For ease of notation, we present the definition in the following form:
\begin{align*}
&\min \quad  \langle \mathbf{C}, \mathbf{T} \rangle  \nonumber\\
& \text{s.t.  } \mathbf{T} \in U(\mu, \nu)= \{\mathbf{T} \in \mathbb{R}^{m \times n}_{+} \mid \mathbf{T}\mathbf{1}_n = \mu, \mathbf{T}^T\mathbf{1}_m = \nu   \},
\end{align*}
where $\mathbf{C}$ denotes the cost matrix and $\langle \mathbf{C}, \mathbf{T} \rangle = \sum_{ij} \mathbf{C}_{ij}\mathbf{T}_{ij} $ is the inner product between matrices.

In this formulation, $\mathbf{c}_{ij}$ represents the cost between point $\mathbf{x}_i$ and point $\mathbf{y}_j$. It could be any non-negative cost function that captures the transportation cost between the points. The objective is to minimize the total cost, which is the sum of the products of the transportation amounts $\mathbf{T}_{ij}$ and their corresponding costs $\mathbf{c}_{ij}$. We denote the optimal transport distance as $\mathcal{W}(\mu, \nu)$.

The constraints ensure that the transportation plan satisfies the conservation of mass: the total mass transported from each point in $\mathbf{X}$ should be equal to its mass in distribution $\mu$, and the total mass received at each point in $\mathbf{Y}$ should be equal to its mass in distribution $\nu$. Additionally, the transportation amounts $\mathbf{T}_{ij}$ are non-negative.

The solution to this optimization problem provides the optimal transportation plan, which specifies how much mass is transported from each point in $\mathbf{X}$ to each point in $\mathbf{Y}$ to minimize the total cost. Algorithms, such as the Hungarian algorithm \cite{kuhn1955hungarian} can be applied to solve this problem with the complexity of $O(m^2n)$.

The computational complexity to solve the general optimal transport problem is relatively high. \citet{cuturi2013sinkhorn} proposes an entropic regularized version of the optimal transport problem as follows:
\begin{align*}
\min \quad & \langle \mathbf{C}, \mathbf{T} \rangle - \epsilon \text{H}(\mathbf{T})  \nonumber\\
\text{subject to}\quad & \mathbf{T} \in U(\mu, \nu),
\end{align*}
where $\epsilon>0$ is a hyperparameter and $\text{H}(\mathbf{T})=\sum_{i, j} (1-\log \mathbf{T}_{ij})\mathbf{T}_{ij}$.

This regularized problem can be solved by the Sinkhorn algorithm efficiently with a complexity of $O(\frac{mn}{\epsilon})$. It can be shown that this regularized version approximately solves the initial discrete optimal transport problem.

\section{Understanding FreeMatch via Optimal Transport} \label{OT understand}

In this section, we start by using the view of semantic matching of distributions to understand the state-of-the-art pseudo-labeling-based semi-supervised learning methods. The main tool we will use is the optimal transport. Without loss of generality, we take the state-of-the-art method FreeMatch \cite{wang2022freematch} as an example. For simplicity, we abbreviate the exponential moving average (EMA) operation.

We find that a natural semantic distribution naturally arises in (semi-) supervised learning, which is the semantic distribution between samples and classes. Specifically, denote $\mu = \frac{1}{m} \mathbf{1}_m$, then the following set of matrices can be seen as the semantic distribution between $m$ samples and $K$ classes (Because each row of $\mathbf{T}$ captures the semantic of the sample's relationship to each class, the bigger the semantic distribution, the bigger the values.):
$$U(\mu)=\{\mathbf{T} \in \mathbb{R}^{m \times K}_{+} \mid \mathbf{T}\mathbf{1}_K = \mu \},$$
where $\mathbf{1}_K$ is an all one vector.

The above observation is also noticed in the setting of supervised learning setting by \citep{shi2023understanding}. In the following, we will first recap the derivation in their paper for supervised loss, where \citep{shi2023understanding} introduces the framework of inverse optimal transport (IOT).

IOT aims to infer the cost function from the observed empirical semantic distribution matrix. It usually parameterizes the cost matrix into a learnable matrix $\mathbf{C}^{\theta}$ and solves the following optimization problem:
\begin{align} \label{IOT}
\min \quad & \text{KL}(\bar{\mathbf{T}} \| \mathbf{T}^{\theta}) \nonumber\\
\text{subject to}\quad & \mathbf{T}^{\theta} = \argmin_{\mathbf{T} \in U(\mu)} \langle \mathbf{C}^{\theta}, \mathbf{T} \rangle - \epsilon \text{H}(\mathbf{T}), 
\end{align}
where $\bar{\mathbf{T}}$ is a given semantic distribution matrix and the KL divergence is defined as follows.

\begin{definition} 
For any two {positive} measures (distributions) $\mathbf{P}$ and $\mathbf{Q}$ on the same support $\mathcal{X}$, the KL divergence from $\mathbf{Q}$ to $\mathbf{P}$ is given by:
\begin{equation} \label{KL divergence}
\text{KL}(\mathbf{P} \| \mathbf{Q}) = - \sum_{x \in \mathcal{X}} \mathbf{P}(x) \log \frac{\mathbf{P}(x)}{\mathbf{Q}(x)} - \sum_{x \in \mathcal{X}} \mathbf{P}(x) + \sum_{x \in \mathcal{X}} \mathbf{Q}(x).    
\end{equation}
\end{definition}

\citet{shi2023understanding} show that 
\begin{equation} \label{closed form transport}
\mathbf{T}_{ij}= \frac{1}{m} \frac{\exp{(-\mathbf{C}_{ij}/\epsilon)}}{\sum^{K}_{k=1} \exp{(-\mathbf{C}_{ik}/\epsilon)}}    
\end{equation}
is the closed-form solution to the optimization problem (\ref{relax OT}):
\begin{equation} \label{relax OT}
\argmin_{\mathbf{T} \in U(\mu)} \quad \langle \mathbf{C}, \mathbf{T} \rangle - \epsilon \text{H}(\mathbf{T}).    
\end{equation}

Then consider a batch of labeled data is $\{(\mathbf{x}_i, \mathbf{y}_i)\}^{B}_{i=1}$, where $\mathbf{x}_i$ represents the $i$-th image in the dataset and $\mathbf{y}_i$ is the label for this image. 
We can construct a ``ground truth'' semantic distribution matrix $\bar{\mathbf{T}}$ by setting $\bar{\mathbf{T}}_{ij}=\frac{1}{B}\delta^{\mathbf{y}_i}_{j}$. 

Denote the logits generated by the neural network for each image $\mathbf{x}_i$ as $\mathbf{l}_{\theta}(x_i)$. By setting the cost matrix $\mathbf{C}^{\theta}_{ij}= c-\mathbf{l}_{\theta}(x_i)_j$ ($c$ is a large constant), simplifying the transport matrix (\ref{closed form transport}) by dividing the same constant $\exp(-c/\epsilon)$, assuming there are a total of $K$ labels, the transportation matrix is given by:
\begin{equation} \label{matching matrix}
\mathbf{T}^{\theta}_{ij}= \frac{1}{m} \frac{\exp(\mathbf{l}_{\theta}(\mathbf{x}_i)_j / \epsilon)}{\sum^{K}_{k=1} \exp(\mathbf{l}_{\theta}(\mathbf{x}_i)_k / \epsilon)}.    
\end{equation}

It is then straightforward to find that the loss in problem ($\ref{IOT}$) is reduced as follows:
\begin{equation*}
\mathcal{L}= - \sum^{B}_{i=1} \log \frac{\exp(\mathbf{l}_{\theta}(x_i)_j / \epsilon)}{\sum^{K}_{k=1} \exp(\mathbf{l}_{\theta}(\mathbf{x}_i)_k / \epsilon)}  + \text{Const,} 
\end{equation*}
which exactly mirrors the supervised cross-entropy loss in semi-supervised learning with a temperature parameter $\epsilon$. 

Next, we delve into comprehending the more challenging unsupervised loss.
We introduce a lemma that is very useful afterward in analyzing the unsupervised loss.
\begin{lemma} \label{mean lemma}
$\frac{\sum^{m}_{i=1} s_i}{m}$ is the unique solution of the optimization problem:
\begin{equation*}
\min_{x } \mathcal{W}(\delta_x, \sum^m_{i=1} \frac{1}{m} \delta_{s_i} ),   
\end{equation*}
where the underlying cost is the square of $l^2$ distance.
\end{lemma}

FreeMatch employs an adaptive threshold for pseudo-labeling.
It is essentially equivalent to generating the threshold based on the semantic distribution matrix of (the teacher) model from equation (\ref{matching matrix}). 
In particular, we start by analyzing the global threshold in FreeMatch, which aims to modulate the global confidence across different classes. 
Given that, each row of the matching matrix indicates the estimated probability $\mathbf{q}_{u_i}$ over the $K$ classes. 
An intuitive idea is to associate each unlabeled sample $u_i$ with a real number indicating the prediction confidence, thus we can take $\text{max}(\mathbf{q}_{u_i})$ as a representative. 
Consequently, the general prediction confidence over the unlabeled data can be represented by a probability distribution $\sum^{\mu B}_{i=1} \frac {1}{\mu B} \delta_{\text{max}(\mathbf{q}_{u_i})}$ that captures the full knowledge of the predictions. 
By identifying the global threshold $\tau$ with a probability distribution $\delta_{\tau}$ and using Lemma \ref{mean lemma}, the global threshold can be calculated as $$\tau = \frac{\sum^{\mu B}_{i=1} \text{max}(\mathbf{q}_{u_i})}{\mu B}.$$

Since the global threshold does not accurately reflect the learning status of each class, we can refine the global threshold by incorporating the learning information for each class.
Note the prediction $\mathbf{q}_{u_i}$ not only provide the ``best'' confidence $\text{max}(\mathbf{q}_{u_i})$, but also suggest the confidence on each class $k$ ($1 \leq k \leq K$). 

By aggregating all confidences of unlabeled data for class $k$ and organizing them into a probability distribution $\sum^{\mu B}_{i=1} \frac {1}{\mu B} \delta_{\mathbf{q}_{u_i}(k)}$, we can calculate the local importance $\mathbf{p}_1(k) = \sum^{\mu B}_{i=1} \frac {\mathbf{q}_{u_i}(k)}{\mu B}$ using a similar argument in the global threshold case.
By adjusting the relative threshold according to the importance, we can finally derive the (local) threshold as follows:
\begin{equation*}
\tau(k) = \frac{\mathbf{p}_1(k)}{\text{max}_{k^{\prime}} \mathbf{p}_1(k^{\prime})}  \frac{\sum^{\mu B}_{i=1} \text{max}(\mathbf{q}_{u_i})}{\mu B}.
\end{equation*}
When dealing with unlabeled data, there is no ``ground truth'' semantic distribution matrix like the supervised cases. Therefore, we use the semantic distribution matrix \textbf{after} threshold filtering to serve as a ``ground truth''.
It's important to highlight that both the teacher and student models share a similar format of the semantic distribution matrix given by equation (\ref{matching matrix}). 
Consequently, unlabeled samples filtered by the teacher model are also excluded from the student model to avoid transferring predictions with low confidence. We use different $\epsilon$ for the teacher and student models (student use $\epsilon=1$) as they have different confidences, when the $\epsilon$ for the teacher model is approaching zero, equation (\ref{matching matrix}) will recover the one-hot pseudo label. 

Thus the reduced semantic distribution matrices of teacher and student models are no longer probability matrices. 
As they still form positive matrices, by converting each row of the teacher model's predictions into pseudo labels and using the definition of KL divergence in equation (\ref{KL divergence}) for positive distributions (measures). We find that the KL divergence between the teacher and student semantic distribution matrices recovers exactly the unsupervised loss $\mathcal{L}_{\text{un1}}$. 
The fairness loss $\mathcal{L}_{\text{un2}}$ can be similarly understood according to Lemma \ref{mean lemma}.

\noindent\textbf{Remark}: More algorithms derived from the framework of using optimal transport to match semantics can be found in Appendix \ref{More algorithms}.

\section{OTMatch: Improving Semi-Supervised Learning with Optimal Transport}

From the derivation in section \ref{OT understand}, we found that the performance of student models depends on the semantic distribution given by the teacher model. As teacher model generates one-hot pseudo labels, which makes it more likely to be over-confident in its predictions. This overconfidence leads to misclassifications and hampers the model's performance. To tackle this issue, we propose a novel solution that incorporates inter-class semantic relationships to alleviate model overconfidence in pseudo-labeling-based methods. Optimal transport is also a suitable tool here because the pseudo-label and the predicted probability given by the student model are two distributions. By considering comprehensive inter-class relationships instead of relying on a single-category target, we aim to improve the model's robustness and accuracy.

\begin{algorithm}[tbp]

\caption{OTMatch training algorithm at $t$-th step}
\label{alg:freematch123}
\begin{algorithmic}[1]
\STATE \textbf{Input:} Number of classes $K$, labeled samples $\{ (\mathbf{x}_{l_i}, \mathbf{y}_{l_i}) \}^{B}_{i=1}$, unlabeled samples $\{ \mathbf{x}_{u_i} \}^{\mu B}_{i=1}$, FreeMatch loss weights $w_1$, $w_2$, and EMA decay $m$, OT loss balancing weight $\lambda$, normalized classification head vectors $\{\mathbf{v}_i \}^K_{i=1}$.

\STATE \textbf{FreeMatch loss:}

\STATE Calculate $\mathcal{L}_{\text{sup}}$ using equation (\ref{L_sup})

\STATE $\tau_t = m \tau_{t-1} + (1-m) \frac{1}{\mu B} \sum_{i=1}^{\mu B} \text{max} (\mathbf{q}_{u_i})$

\STATE $\tilde{p}_t = m \tilde{p}_{t-1} + (1-m) \frac{1}{\mu B}\sum_{b=1}^{\mu B} \mathbf{q}_{u_i}$

\STATE $\tilde{h}_t = m \tilde{h}_{t-1} + (1 - m) \operatorname{Hist}_{\mu B}  \left( \hat{\mathbf{q}}_{u_i} \right)$

\FOR{$c=1$ to $K$}
\STATE $\tau_t(c) = \operatorname{MaxNorm}(\tilde{p}_t(c)) \cdot \tau_t$
\ENDFOR 

\STATE Calculate $\mathcal{L}_{\text{un1}}$ using equation (\ref{L_un1}) 

\STATE $\overline{p} = \frac{1}{\mu B} \sum_{i=1}^{\mu B} \mathbf{I}\left(\text{max} \left(\mathbf{q}_{u_i}\right) \geq \tau_t(\argmax \left(\mathbf{q}_{u_i}\right)  \right) \mathbf{Q}_{u_i}$ 

\STATE $\overline{h} = \operatorname{Hist}_{\mu B} ( \mathbf{I}\left(\max \left(\textbf{q}_{u_i}\right) \geq \tau_t(\argmax \left(\textbf{q}_{u_i}\right) \right) \hat{\textbf{Q}}_{u_i} ))$

\STATE Calculate $\mathcal{L}_{\text{un2}}$ using equation (\ref{L_un2})

\STATE $\mathcal{L}_{\text{FreeMatch}} = \mathcal{L}_{\text{sup}} + w_1 \mathcal{L}_{\text{un1}} + w_2 \mathcal{L}_{\text{un2}} $

\STATE \textbf{Cost update:}\\
$\mathbf{C}_t(i, j) = m \mathbf{C}_{t-1}(i, j) + (1-m)(1- \langle \mathbf{v}_i, \mathbf{v}_j \rangle)$

\STATE \textbf{OT loss:}\\
$\mathcal{L}_{\text{un3}} = \frac{1}{\mu B} \sum^{\mu B}_{i=1} \mathbf{I}(\max(\mathbf{q}_{u_i})> \tau_t(\argmax(\mathbf{q}_{u_i})))$ \\
\quad \quad \quad $\sum^{K}_{k=1} \mathbf{C}_{t}(\argmax(\mathbf{q}_{u_i}), k)\mathbf{Q}_{u_i}(k)$

\STATE \textbf{OTMatch loss:}\\
$\mathcal{L}_{\text{OTMatch}} = \mathcal{L}_{\text{FreeMatch}} + \lambda \mathcal{L}_{un3}$
\end{algorithmic}
\end{algorithm}

\begin{table*}
\centering
\caption{Error rates (100\% - accuracy) on CIFAR-10/100, and STL-10 datasets for state-of-the-art methods in semi-supervised learning. Bold indicates the best performance, and underline indicates the second best.}
\label{tab:results-semi}
\small
\resizebox{\textwidth}{!}{%
\begin{tabular}{l|ccc|cc|cc}
\toprule
Dataset & \multicolumn{3}{c|}{CIFAR-10} & \multicolumn{2}{c|}{CIFAR-100} & \multicolumn{2}{c}{STL-10} \\ \cmidrule{1-1}\cmidrule(lr){2-4}\cmidrule(lr){5-6}\cmidrule{7-8} 
\# Label & 10 & 40 & 250 & 400 & 2500 & 40 & 1000\\ \cmidrule{1-1}\cmidrule(lr){2-4}\cmidrule(lr){5-6}\cmidrule{7-8}
$\Pi$ Model \cite{rasmus2015semi} & 
79.18{\scriptsize $\pm$1.11} &
74.34{\scriptsize $\pm$1.76} & 46.24{\scriptsize $\pm$1.29} & 86.96{\scriptsize $\pm$0.80} & 58.80{\scriptsize $\pm$0.66} & 74.31{\scriptsize $\pm$0.85} & 32.78{\scriptsize $\pm$0.40} \\
Pseudo Label \cite{lee2013pseudo} & 80.21{\scriptsize $\pm$ 0.55} & 74.61{\scriptsize $\pm$0.26} & 46.49{\scriptsize $\pm$2.20} & 87.45{\scriptsize $\pm$0.85} & 57.74{\scriptsize $\pm$0.28} & 74.68{\scriptsize $\pm$0.99} & 32.64{\scriptsize $\pm$0.71} \\
VAT \cite{miyato2018virtual} & 79.81{\scriptsize $\pm$ 1.17} & 74.66{\scriptsize $\pm$2.12} & 41.03{\scriptsize $\pm$1.79} & 85.20{\scriptsize $\pm$1.40} & 46.84{\scriptsize $\pm$0.79} & 74.74{\scriptsize $\pm$0.38} & 37.95{\scriptsize $\pm$1.12} \\
MeanTeacher \cite{tarvainen2017mean} & 76.37{\scriptsize $\pm$ 0.44} & 70.09{\scriptsize $\pm$1.60} & 37.46{\scriptsize $\pm$3.30} & 81.11{\scriptsize $\pm$1.44} & 45.17{\scriptsize $\pm$1.06} & 71.72{\scriptsize $\pm$1.45} & 33.90{\scriptsize $\pm$1.37} \\
MixMatch \cite{berthelot2019mixmatch} & 65.76{\scriptsize $\pm$ 7.06} & 36.19{\scriptsize $\pm$6.48} & 13.63{\scriptsize $\pm$0.59} & 67.59{\scriptsize $\pm$0.66} & 39.76{\scriptsize $\pm$0.48} & 54.93{\scriptsize $\pm$0.96} & 21.70{\scriptsize $\pm$0.68} \\
ReMixMatch  \cite{berthelot2019remixmatch} & 20.77{\scriptsize $\pm$ 7.48} & 9.88{\scriptsize $\pm$1.03} & 6.30{\scriptsize $\pm$0.05} & 42.75{\scriptsize $\pm$1.05} & \textbf{26.03{\scriptsize $\pm$0.35}} & 32.12{\scriptsize $\pm$6.24} & 6.74{\scriptsize $\pm$0.17}\\
UDA \cite{xie2020unsupervised} & 34.53{\scriptsize $\pm$ 10.69} & 10.62{\scriptsize $\pm$3.75} & 5.16{\scriptsize $\pm$0.06} & 46.39{\scriptsize $\pm$1.59} & 27.73{\scriptsize $\pm$0.21} & 37.42{\scriptsize $\pm$8.44} & 6.64{\scriptsize $\pm$0.17} \\
FixMatch \cite{sohn2020fixmatch} & 24.79{\scriptsize $\pm$ 7.65} & 7.47{\scriptsize $\pm$0.28} & 5.07{\scriptsize $\pm$0.05} & 46.42{\scriptsize $\pm$0.82} & 28.03{\scriptsize $\pm$0.16} & 35.97{\scriptsize $\pm$4.14} & 6.25{\scriptsize $\pm$0.33} \\
Dash \cite{xu2021dash} & 27.28{\scriptsize $\pm$ 14.09} & 8.93{\scriptsize $\pm$3.11} & 5.16{\scriptsize $\pm$0.23} & 44.82{\scriptsize $\pm$0.96} & 27.15{\scriptsize $\pm$0.22} & 34.52{\scriptsize $\pm$4.30} & 6.39{\scriptsize $\pm$0.56} \\
MPL \cite{pham2021meta} & 23.55{\scriptsize $\pm$ 6.01} & 6.93{\scriptsize $\pm$0.17} & 5.76{\scriptsize $\pm$0.24} & 46.26{\scriptsize $\pm$1.84} & 27.71{\scriptsize $\pm$0.19} & 35.76{\scriptsize $\pm$4.83} & 6.66{\scriptsize $\pm$0.00} \\

FlexMatch \cite{zhang2021flexmatch} & 13.85{\scriptsize $\pm$ 12.04} & 4.97{\scriptsize $\pm$0.06} & 4.98{\scriptsize $\pm$0.09} & 39.94{\scriptsize $\pm$1.62} & 26.49{\scriptsize $\pm$0.20} & 29.15{\scriptsize $\pm$4.16} & 5.77{\scriptsize $\pm$0.18} \\
FreeMatch \cite{wang2022freematch} & \underline{8.07{\scriptsize $\pm$ 4.24}} & \underline{4.90{\scriptsize $\pm$0.04}} & \underline{4.88{\scriptsize $\pm$0.18}} & \underline{37.98{\scriptsize $\pm$0.42}} & 
{26.47{\scriptsize $\pm$0.20}} &
\underline{15.56{\scriptsize $\pm$0.55}} & \underline{5.63{\scriptsize $\pm$0.15}} \\
\cmidrule{1-1}\cmidrule(lr){2-4}\cmidrule(lr){5-5}\cmidrule{6-8}
OTMatch (Ours) & \textbf{4.89{\scriptsize $\pm$ 0.76}} & \textbf{4.72{\scriptsize $\pm$ 0.08}} & \textbf{4.60{\scriptsize $\pm$ 0.15}} & \textbf{37.29{\scriptsize $\pm$ 0.76}}  & 
\underline{26.04{\scriptsize $\pm$ 0.21}}  & 
\textbf{12.10{\scriptsize $\pm$ 0.72}} & \textbf{5.60{\scriptsize $\pm$ 0.14}} \\
\bottomrule
\end{tabular}%
}
\end{table*}

The cost function in optimal transport plays an important role. 
\citet{frogner2015learning} construct cost using additional knowledge like word embedding. This approach may not be problem-specific and incorporates additional knowledge. Unlike previous works, we propose that the cost matrix can actually be effectively bootstrapped from the model itself. The basic idea is to ``infer'' the cost from the learning dynamic of the model. Since the model parameters are updated from (the stochastic) gradient descent method, our initial step involves analyzing the gradients. To simplify the analysis, assume the feature extracted by the model as unconstrained variables. 
Suppose the last layer of the neural network weights are denoted by $\mathbf{W}=[\mathbf{w}_1 \mathbf{w}_2 \cdots \mathbf{w}_K]$.
Define the predicted probability matching for the image $\mathbf{x}$ as follows:
\begin{equation*}
\mathbf{p}_k(f_{\theta}(\mathbf{x}))=\frac{\exp \left(f_{\theta}(\mathbf{x})^T \mathbf{w}_k\right)}{\sum_{k^{\prime}=1}^K \exp \left(f_{\theta}(\mathbf{x})^T \mathbf{w}_{k^{\prime}}\right)}, 1 \leq k \leq K.    
\end{equation*}
Then we can calculate the loss's gradient for each image embedding with a label (or pseudo label) $k$ as follows:
\begin{equation*}
\frac{\partial \mathcal{L}}{\partial f_{\theta}(\mathbf{x})}=-\left(1-\mathbf{p}_k(f_{\theta}(\mathbf{x}))\right) \mathbf{w}_k+\sum_{k^{\prime} \neq k}^K \mathbf{p}_{k^{\prime}}(f_{\theta}(\mathbf{x})) \mathbf{w}_{k^{\prime}},    
\end{equation*}
where $\mathcal{L}$ is the supervised loss $\mathcal{L}_{sup}$ or unsupervised loss $\mathcal{L}_{un1}$.

As the goal is to push $f_{\theta}(\mathbf{x})$ to the direction of $\mathbf{w}_k$, the updated score $U(\mathbf{x})$ along $\mathbf{w}_k$ during SGD on $f_{\theta}(\mathbf{x})$ can be calculated as:
\begin{align} \label{gradient update formula}
&U(\mathbf{x}) = \langle-\frac{\partial \mathcal{L}}{\partial f_{\theta}(\mathbf{x})}, \mathbf{w}_k \rangle \nonumber \\
& = \left(1-\mathbf{p}_k(f_{\theta}(\mathbf{x}))\right) \langle \mathbf{w}_k, \mathbf{w}_k \rangle - \sum_{k^{\prime} \neq k}^K \mathbf{p}_{k^{\prime}}(f_{\theta}(\mathbf{x})) \langle \mathbf{w}_{k^{\prime}}, \mathbf{w}_k \rangle.   
\end{align}
Note $U(\mathbf{x})$ reflects the hardness of classifying image $\mathbf{x}$ into class $k$.
Thus we would like our expected cost of classification $C(\mathbf{x})$ to be proportional to $U(\mathbf{x})$. Using the law of probability, we can decompose $C(\mathbf{x})$ as follows:
\begin{equation*}
C(\mathbf{x}) = \mathbb{E}_k(\text{Cost} \mid \mathbf{x}) = \sum^{K}_{k^{\prime} =1} \mathbf{C}_{kk^{\prime}} p_{k^{\prime}}(f_{\theta}(\mathbf{x})). 
\end{equation*}
When $\| \mathbf{w}_i \|_2=1$, by setting $\mathbf{C}_{kk^{\prime}}= 1-\langle \mathbf{w}_k, \mathbf{w}_{k^{\prime}} \rangle$ we can demonstrate that $C(\mathbf{x}) = U(\mathbf{x})$.

Hence, it is evident that inter-class semantic relationships are indeed the core of the construction of an effective cost.
Taking the fluctuation of batch training into consideration, we finally derive the cost update formula as follows:
\begin{equation*}
\mathbf{C}_{kk^{\prime}} = m \mathbf{C}_{kk^{\prime}} + (1-m) (1 - \langle \mathbf{v}_k, \mathbf{v}_{k^{\prime}} \rangle),    
\end{equation*}
where the cost is initialized by discrete metric, $m$ is the momentum coefficient, and $\mathbf{v}_k= \frac{\mathbf{w}_k}{\| \mathbf{w}_k \|_2}$.

The computation cost of calculating the optimal transport cost is relatively high, which hinders its application. 
We present a lemma that shows under some mild conditions in semi-supervised learning, optimal transport can be calculated in complexity $O(K)$.
\begin{lemma} \label{W dis}
Suppose two probability distributions $\mu$ and $\nu$ support on $\mathcal{X}$ and suppose $\mid \mathcal{X} \mid=K$. Suppose the cost is generated by a metric and there exists $k$ such that $\mu(i) \leq \nu(i)$ for any $i \neq k$. Then $\mathcal{W}(\mu, \nu) = \sum^{K}_{i=1} \mathbf{C}_{ik} (\nu(i)-\mu(i))$.   
\end{lemma}
Thus we can finally obtain our optimal transport-based unsupervised loss as follows:
\begin{equation*}
\mathcal{L}_{\text{un3}} = \frac{1}{\mu B} \sum^{\mu B}_{i=1} \mathbf{I}(\max(\mathbf{q}_{u_i})> \tau(\argmax(\mathbf{q}_{u_i}))) \mathbf{W}(\hat{\mathbf{q}}_{u_i}, \mathbf{Q}_{u_i}).
\end{equation*}

When combining our method with FreeMatch,
we obtain our final OTMatch loss as $\mathcal{L}_{\text{OTMatch}}= \mathcal{L}_{\text{FreeMatch}}+\lambda \mathcal{L}_{\text{un3}}$, where $\lambda$ is a hyperparameter. This loss considers incorporating semantic information to distinguish two distribution $\hat{\mathbf{q}}_{u_i}$ and $\mathbf{Q}_{u_i}$.
The whole process of our method is outlined in Algorithm \ref{alg:freematch123}.

Interestingly, the loss $\mathcal{L}_{un3}$ can also be interpreted using the view of self-attention \citep{vaswani2017attention}. 
Setting $f_{\theta}(\mathbf{x})$ as query, $\mathbf{w}_j$ ($1 \leq j \leq K$) as keys and $\mathbf{v}_j$ ($1 \leq j \leq K$) as values, recall the definition of self-attention, $\sum^{K}_{i=1} \mathbf{p}_i(f_{\theta}(\mathbf{x})) \mathbf{v}_i$ is exactly the representation generated by self-attention. 

For an unlabeled image $\mathbf{x}$ with pseudo label $k$, the loss can be reformulated as: $\mathcal{W}(\delta_k, \text{Pr}(\mathbf{x}))  = \sum^{K}_{i=1} \mathbf{C}_{ik} \text{Pr}(i \mid \mathbf{x})  
= \sum^{K}_{i=1} (1- \langle \mathbf{v}_i, \mathbf{v}_k \rangle) \mathbf{p}_i(f_{\theta}(\mathbf{x})) 
= 1- \langle \sum^{K}_{i=1} \mathbf{p}_i(f_{\theta}(\mathbf{x})) \mathbf{v}_i, \mathbf{v}_k \rangle.$ 
Thus intuitively, the loss $\mathcal{L}_{\text{un3}}$ seeks to align the representation generated by the self-attention mechanism with the classification head vector $\mathbf{v}_k$.

\section{Experiments}
\begin{table}[t]
\centering
\caption{Error rates (100\% - accuracy) on ImageNet with 100 labels per class.}
\label{tab:ImageNet}
\small
\begin{tabular}{l|cc}
\toprule
               & Top-1 & Top-5 \\ \midrule
FixMatch \citep{sohn2020fixmatch}       & 43.66 & 21.80 \\
FlexMatch \citep{zhang2021flexmatch}      & 41.85 & 19.48 \\
FreeMatch \citep{wang2022freematch}      & 40.57 & 18.77 \\
OTMatch (Ours) & \textbf{39.29} & \textbf{17.77} \\ \bottomrule
\end{tabular}
\end{table}

\begin{table}[]
\renewcommand{\arraystretch}{1}
\setlength{\tabcolsep}{4pt}
\centering
\caption{Comparisons with state-of-the-art semi-supervised learning methods on Amazon Review and Yelp Review. Error rates (100\% - accuracy) are reported.}
\label{nlp-results}
\small
\begin{tabular}{ccccc}
\toprule
\textbf{}           & \multicolumn{2}{c}{\textbf{Amz Review}}                                  & \multicolumn{2}{c}{\textbf{Yelp Review}}                           \\
\textbf{\# Label}     & \multicolumn{1}{c}{250} & \multicolumn{1}{c}{1000} & \multicolumn{1}{c}{250} & \multicolumn{1}{c}{1000} \\ \midrule
FixMatch~\citep{sohn2020fixmatch}           & 47.85                                    & 43.73                            & 50.34                           & 41.99                            \\
CoMatch~\citep{li2021comatch}             & 48.98                                    & 44.37                            & 46.49                           & 41.11                            \\
Dash~\citep{xu2021dash}             & 47.79                                    & 43.52                            & 35.10                           & 30.51                            \\
AdaMatch~\citep{adamatch}            & 46.75                                    & 43.50                            & 48.16                           & 41.71                            \\
SimMatch~\citep{zheng2022simmatch}            & 47.27                                    & 43.09                            & 46.40                           & 41.24                            \\ \midrule
FlexMatch~\citep{zhang2021flexmatch}           & 45.75                                    & 43.14                            & 46.37                           & 40.86                            \\
OTMatch (Ours) & \textbf{43.81}                           & \textbf{42.35}                   & \textbf{43.61}                  & \textbf{39.76}                   \\ \bottomrule
\end{tabular}
\end{table}

\subsection{Setup}

Based on previous studies \citep{sohn2020fixmatch,zhang2021flexmatch,wang2022freematch}, we evaluate our method on widely used vision semi-supervised benchmark datasets, including CIFAR-10/100, STL-10, and ImageNet. Our approach (OTMatch) incorporates the optimal transport loss with the calculation of the unsupervised loss within FreeMatch. Our experiments primarily focus on realistic scenarios with limited labeled data. We utilize SGD as the optimizer with a momentum of 0.9 and a weight decay of $5\times 10^{-4}$. The learning rate follows a cosine annealing scheduler, initialized at $0.03$. The batch size is set to 64, except for ImageNet where it is 128. The ratio of unlabeled data to labeled data is 7. We report results over multiple runs over seeds. Regarding the choice of backbones, we use the Wide ResNet28-2 for CIFAR-10, Wide ResNet-28-8 for CIFAR-100, Wide ResNet-37-2 for STL-10, and ResNet-50 for ImageNet. Our training process consists of $2^{20}$ total training iterations, where each step involves sampling an equal number of labeled images from all classes. For the hyperparameter settings of our method, we set $\lambda=0.5$ for CIFAR-10, $\lambda=0.15$ for STL-10 and CIFAR-100, and $\lambda=0.01$ for ImageNet. The momentum coefficient of the cost update is set to $0.999$.

\begin{table}[t]
\renewcommand{\arraystretch}{1}
\setlength{\tabcolsep}{4pt}
\centering
\caption{Ablation studies on the chosen cost in the optimal transport loss. Error rates (100\% - accuracy) on CIFAR-10 with 4 labels per class are reported.}
\label{tab:cost ablation}
\small
\begin{tabular}{l|c} 
\toprule
               & Top-1 \\ \midrule
Binary Cost       & 5.20  \\
Cost Based on Covariance      & 4.88 \\
OTMatch Cost (Ours) & \textbf{4.72}   \\ \bottomrule
\end{tabular}
\end{table}

\begin{figure}[t]
\centering 
\includegraphics[width = 8cm, height=6cm]{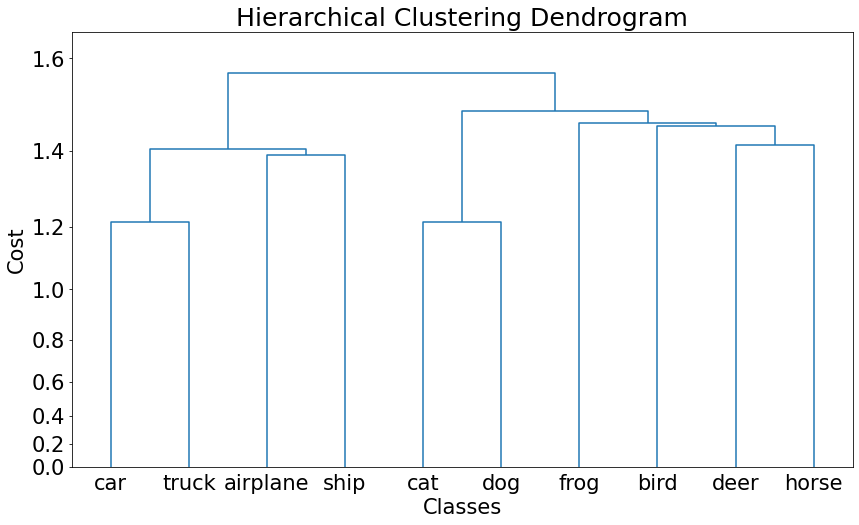}
\caption{Hierarchical clustering results of the learned cost matrix on CIFAR-10.}
\label{fig: cluster}
\end{figure}

\subsection{Results}
\paragraph{Performance improvements.}In our evaluation, we compare our approach to a wide range of representative semi-supervised learning methods, including $\Pi$-Model \citep{rasmus2015semi}, Pseudo-Label \citep{lee2013pseudo}, VAT \citep{miyato2018virtual}, MeanTeacher  \citep{tarvainen2017mean}, MixMatch \citep{berthelot2019mixmatch}, ReMixMatch \citep{berthelot2019remixmatch}, UDA \citep{xie2020unsupervised}, Dash  \citep{xu2021dash}, MPL \citep{pham2021meta}, FixMatch \citep{sohn2020fixmatch}, FlexMatch \citep{zhang2021flexmatch}, and FreeMatch \citep{wang2022freematch}. The results are reported in Table~\ref{tab:results-semi} and \ref{tab:ImageNet}.

It's evident that OTMatch outperforms previous methods across the board and notably enhances performance.
This improvement is particularly pronounced in scenarios with limited labeled samples such as STL-10 with 40 labels and CIFAR-10 with 10 labels, which indeed aligns with our motivation.
It is important to note that in CIFAR-10 cases, fully supervised has achieved an error rate of $4.62$ \citep{wang2022freematch}. 
Thus, our method further closes the gap between semi-supervised learning and fully supervised learning.

Furthermore, as optimal transport can be incorporated wherever cross-entropy is used, our method can seamlessly integrate with recent and future techniques thereby achieving greater performance enhancements.
The computational complexity is only $O(K)$, making it computationally friendly even as the number of labels scales. This highlights optimal transport as a useful regularizer with minimal computation overhead.
 
\subsection{Analysis}
\noindent\textbf{Results on data from other modalities.}
To demonstrate the utility of our approach, we further extend our evaluations to encompass USB datasets \citep{usb} of language modality. 
Specifically, the results in Table \ref{nlp-results} demonstrate that on both Amazon Review and Yelp Review, when our approach is integrated with Flex-Match (current state-of-the-art), we have achieved an improvement, reaching a new state-of-the-art. 
This also validates the fact that, beyond the compatibility with FreeMatch, our approach actually can be effortlessly integrated with various existing methods. 

\noindent\textbf{Ablations of the cost.}
The cost plays a crucial role in the optimal transport loss, we conduct further ablation studies to investigate its effect. The binary cost $\mathbf{c}_1(x,y)= \mathbb{I}_{x \neq y}$ is a straightforward cost option. 
However, it does not take into account the relationships between classes.
Additionally, we explore an alternative cost formulation that considers class relationships. 
In this regard, we update the cost based on the covariance matrix of predicted probabilities for strongly augmented images.
We compare the performance of these costs on CIFAR-10 with 40 labels benchmark and summarize the results in Table \ref{tab:cost ablation}. 
It is clear that the cost used in our method achieves the best performance.

In addition, we also demonstrate the hierarchical clustering results of the final cost matrix in Fig. \ref{fig: cluster}, revealing that correct inter-class semantic relationships are learned. Specifically, In the hierarchical clustering, we can observe that various classes that denote non-living things such as airplane, truck, and ship are clustered closely together, while classes that denote living things (animals) like cat, dog, frog, bird, deer, and horse are also clustered closely together. More interestingly, we can also observe more fine-grained clustering effects, such as the proximity between car and truck, cat and dog, as well as deer and horse.

\section{Conclusion}
In this paper, we present a fresh perspective for semi-supervised learning, going beyond the previous efforts on solely improving the quality of pseudo-labels. 
We introduce a novel algorithm, OTMatch, that harnesses the inherent relationships between classes with inverse optimal transport.
We also demonstrate the superiority of OTMatch in our experiments.

By introducing OTMatch, we not only contribute to the advancement of semi-supervised learning techniques but also pave the way for future research by promoting the incorporation of optimal transport loss in a versatile manner.

\section*{Acknowledgment}
Weiran Huang is supported by 2023 CCF-Baidu Open Fund and Microsoft Research Asia.

We would also like to express our sincere gratitude to the reviewers of ICML 2024 for their insightful and constructive feedback. Their valuable comments have greatly contributed to improving the quality of our work.

\bibliography{reference}

\begin{thebibliography}{57}
\providecommand{\natexlab}[1]{#1}
\providecommand{\url}[1]{\texttt{#1}}
\expandafter\ifx\csname urlstyle\endcsname\relax
  \providecommand{\doi}[1]{doi: #1}\else
  \providecommand{\doi}{doi: \begingroup \urlstyle{rm}\Url}\fi

\bibitem[Assran et~al.(2021)Assran, Caron, Misra, Bojanowski, Joulin, Ballas,
  and Rabbat]{assran2021semi}
Assran, M., Caron, M., Misra, I., Bojanowski, P., Joulin, A., Ballas, N., and
  Rabbat, M.
\newblock Semi-supervised learning of visual features by non-parametrically
  predicting view assignments with support samples.
\newblock In \emph{Proceedings of the IEEE/CVF International Conference on
  Computer Vision}, pp.\  8443--8452, 2021.

\bibitem[Berthelot et~al.(2019{\natexlab{a}})Berthelot, Carlini, Cubuk,
  Kurakin, Sohn, Zhang, and Raffel]{berthelot2019remixmatch}
Berthelot, D., Carlini, N., Cubuk, E.~D., Kurakin, A., Sohn, K., Zhang, H., and
  Raffel, C.
\newblock Remixmatch: Semi-supervised learning with distribution alignment and
  augmentation anchoring.
\newblock \emph{arXiv preprint arXiv:1911.09785}, 2019{\natexlab{a}}.

\bibitem[Berthelot et~al.(2019{\natexlab{b}})Berthelot, Carlini, Goodfellow,
  Papernot, Oliver, and Raffel]{berthelot2019mixmatch}
Berthelot, D., Carlini, N., Goodfellow, I., Papernot, N., Oliver, A., and
  Raffel, C.~A.
\newblock Mixmatch: A holistic approach to semi-supervised learning.
\newblock \emph{Advances in neural information processing systems}, 32,
  2019{\natexlab{b}}.

\bibitem[Berthelot et~al.(2022)Berthelot, Roelofs, Sohn, Carlini, and
  Kurakin]{adamatch}
Berthelot, D., Roelofs, R., Sohn, K., Carlini, N., and Kurakin, A.
\newblock Adamatch: {A} unified approach to semi-supervised learning and domain
  adaptation.
\newblock In \emph{ICLR}, 2022.

\bibitem[Caron et~al.(2020)Caron, Misra, Mairal, Goyal, Bojanowski, and
  Joulin]{caron2020unsupervised}
Caron, M., Misra, I., Mairal, J., Goyal, P., Bojanowski, P., and Joulin, A.
\newblock Unsupervised learning of visual features by contrasting cluster
  assignments.
\newblock \emph{Advances in neural information processing systems},
  33:\penalty0 9912--9924, 2020.

\bibitem[Caron et~al.(2021{\natexlab{a}})Caron, Touvron, Misra, J{\'e}gou,
  Mairal, Bojanowski, and Joulin]{caron2021emerging}
Caron, M., Touvron, H., Misra, I., J{\'e}gou, H., Mairal, J., Bojanowski, P.,
  and Joulin, A.
\newblock Emerging properties in self-supervised vision transformers.
\newblock In \emph{Proceedings of the IEEE/CVF International Conference on
  Computer Vision}, pp.\  9650--9660, 2021{\natexlab{a}}.

\bibitem[Caron et~al.(2021{\natexlab{b}})Caron, Touvron, Misra, J\'egou,
  Mairal, Bojanowski, and Joulin]{dino}
Caron, M., Touvron, H., Misra, I., J\'egou, H., Mairal, J., Bojanowski, P., and
  Joulin, A.
\newblock Emerging properties in self-supervised vision transformers.
\newblock In \emph{Proceedings of the International Conference on Computer
  Vision (ICCV)}, 2021{\natexlab{b}}.

\bibitem[Chen et~al.(2023{\natexlab{a}})Chen, Tao, Fan, Wang, Wang, Schiele,
  Xie, Raj, and Savvides]{chen2023softmatch}
Chen, H., Tao, R., Fan, Y., Wang, Y., Wang, J., Schiele, B., Xie, X., Raj, B.,
  and Savvides, M.
\newblock Softmatch: Addressing the quantity-quality trade-off in
  semi-supervised learning.
\newblock \emph{International Conference on Learning Representations (ICLR)},
  2023{\natexlab{a}}.

\bibitem[Chen et~al.(2022)Chen, Du, Zhang, Qian, and Wang]{overconfidence}
Chen, M., Du, Y., Zhang, Y., Qian, S., and Wang, C.
\newblock Semi-supervised learning with multi-head co-training.
\newblock In \emph{AAAI}, pp.\  6278--6286, 2022.

\bibitem[Chen et~al.(2020{\natexlab{a}})Chen, Kornblith, Norouzi, and
  Hinton]{chen2020simple}
Chen, T., Kornblith, S., Norouzi, M., and Hinton, G.
\newblock A simple framework for contrastive learning of visual
  representations.
\newblock \emph{arXiv preprint arXiv:2002.05709}, 2020{\natexlab{a}}.

\bibitem[Chen et~al.(2020{\natexlab{b}})Chen, Kornblith, Swersky, Norouzi, and
  Hinton]{chen2020big}
Chen, T., Kornblith, S., Swersky, K., Norouzi, M., and Hinton, G.~E.
\newblock Big self-supervised models are strong semi-supervised learners.
\newblock \emph{Advances in neural information processing systems},
  33:\penalty0 22243--22255, 2020{\natexlab{b}}.

\bibitem[Chen \& He(2021)Chen and He]{chen2021exploring}
Chen, X. and He, K.
\newblock Exploring simple siamese representation learning.
\newblock In \emph{Proceedings of the IEEE/CVF conference on computer vision
  and pattern recognition}, pp.\  15750--15758, 2021.

\bibitem[Chen et~al.(2023{\natexlab{b}})Chen, Tan, Zhao, Chen, Song, Liang, and
  Lu]{chen2023boosting}
Chen, Y., Tan, X., Zhao, B., Chen, Z., Song, R., Liang, J., and Lu, X.
\newblock Boosting semi-supervised learning by exploiting all unlabeled data.
\newblock In \emph{Proceedings of the IEEE/CVF Conference on Computer Vision
  and Pattern Recognition}, pp.\  7548--7557, 2023{\natexlab{b}}.

\bibitem[Cuturi(2013)]{cuturi2013sinkhorn}
Cuturi, M.
\newblock Sinkhorn distances: Lightspeed computation of optimal transport.
\newblock \emph{Advances in neural information processing systems}, 26, 2013.

\bibitem[Frogner et~al.(2015)Frogner, Zhang, Mobahi, Araya, and
  Poggio]{frogner2015learning}
Frogner, C., Zhang, C., Mobahi, H., Araya, M., and Poggio, T.~A.
\newblock Learning with a wasserstein loss.
\newblock \emph{Advances in neural information processing systems}, 28, 2015.

\bibitem[Gong et~al.(2021)Gong, Wang, and Liu]{gong2021alphamatch}
Gong, C., Wang, D., and Liu, Q.
\newblock Alphamatch: Improving consistency for semi-supervised learning with
  alpha-divergence.
\newblock In \emph{Proceedings of the IEEE/CVF Conference on Computer Vision
  and Pattern Recognition}, pp.\  13683--13692, 2021.

\bibitem[Grill et~al.(2020)Grill, Strub, Altch{\'e}, Tallec, Richemond,
  Buchatskaya, Doersch, Avila~Pires, Guo, Gheshlaghi~Azar,
  et~al.]{grill2020bootstrap}
Grill, J.-B., Strub, F., Altch{\'e}, F., Tallec, C., Richemond, P.,
  Buchatskaya, E., Doersch, C., Avila~Pires, B., Guo, Z., Gheshlaghi~Azar, M.,
  et~al.
\newblock Bootstrap your own latent-a new approach to self-supervised learning.
\newblock \emph{Advances in Neural Information Processing Systems},
  33:\penalty0 21271--21284, 2020.

\bibitem[He et~al.(2019)He, Fan, Wu, Xie, and Girshick]{he2019momentum}
He, K., Fan, H., Wu, Y., Xie, S., and Girshick, R.
\newblock Momentum contrast for unsupervised visual representation learning.
\newblock \emph{arXiv preprint arXiv:1911.05722}, 2019.

\bibitem[Hu et~al.(2021)Hu, Yang, Hu, and Nevatia]{hu2021simple}
Hu, Z., Yang, Z., Hu, X., and Nevatia, R.
\newblock Simple: similar pseudo label exploitation for semi-supervised
  classification.
\newblock In \emph{Proceedings of the IEEE/CVF Conference on Computer Vision
  and Pattern Recognition}, pp.\  15099--15108, 2021.

\bibitem[Huang et~al.(2023)Huang, Shen, Yu, Han, and Liu]{huang2023flatmatch}
Huang, Z., Shen, L., Yu, J., Han, B., and Liu, T.
\newblock Flatmatch: Bridging labeled data and unlabeled data with
  cross-sharpness for semi-supervised learning.
\newblock \emph{Advances in Neural Information Processing Systems},
  36:\penalty0 18474--18494, 2023.

\bibitem[Kantorovich(1942)]{kantorovich1942transfer}
Kantorovich, L.
\newblock On the transfer of masses (in russian).
\newblock In \emph{Doklady Akademii Nauk}, volume~37, pp.\  227, 1942.

\bibitem[Khosla et~al.(2020)Khosla, Teterwak, Wang, Sarna, Tian, Isola,
  Maschinot, Liu, and Krishnan]{khosla2020supervised}
Khosla, P., Teterwak, P., Wang, C., Sarna, A., Tian, Y., Isola, P., Maschinot,
  A., Liu, C., and Krishnan, D.
\newblock Supervised contrastive learning.
\newblock \emph{Advances in Neural Information Processing Systems},
  33:\penalty0 18661--18673, 2020.

\bibitem[Kuhn(1955)]{kuhn1955hungarian}
Kuhn, H.~W.
\newblock The hungarian method for the assignment problem.
\newblock \emph{Naval research logistics quarterly}, 2\penalty0 (1-2):\penalty0
  83--97, 1955.

\bibitem[Laine \& Aila(2016)Laine and Aila]{laine2016temporal}
Laine, S. and Aila, T.
\newblock Temporal ensembling for semi-supervised learning.
\newblock \emph{arXiv preprint arXiv:1610.02242}, 2016.

\bibitem[Lee et~al.(2013)]{lee2013pseudo}
Lee, D.-H. et~al.
\newblock Pseudo-label: The simple and efficient semi-supervised learning
  method for deep neural networks.
\newblock In \emph{Workshop on challenges in representation learning, ICML},
  volume~3, pp.\  896, 2013.

\bibitem[Li et~al.(2021)Li, Xiong, and Hoi]{li2021comatch}
Li, J., Xiong, C., and Hoi, S.~C.
\newblock Comatch: Semi-supervised learning with contrastive graph
  regularization.
\newblock In \emph{Proceedings of the IEEE/CVF International Conference on
  Computer Vision}, pp.\  9475--9484, 2021.

\bibitem[Li et~al.(2022)Li, Chen, He, Xu, Yang, Cao, and Huang]{li2022maxmatch}
Li, Y. J.~X., Chen, Y., He, Y., Xu, Q., Yang, Z., Cao, X., and Huang, Q.
\newblock Maxmatch: Semi-supervised learning with worst-case consistency.
\newblock \emph{IEEE Transactions on Pattern Analysis and Machine
  Intelligence}, 2022.

\bibitem[Miyato et~al.(2018)Miyato, Maeda, Koyama, and
  Ishii]{miyato2018virtual}
Miyato, T., Maeda, S.-i., Koyama, M., and Ishii, S.
\newblock Virtual adversarial training: a regularization method for supervised
  and semi-supervised learning.
\newblock \emph{IEEE transactions on pattern analysis and machine
  intelligence}, 41\penalty0 (8):\penalty0 1979--1993, 2018.

\bibitem[Nassar et~al.(2021)Nassar, Herath, Abbasnejad, Buntine, and
  Haffari]{nassar2021all}
Nassar, I., Herath, S., Abbasnejad, E., Buntine, W., and Haffari, G.
\newblock All labels are not created equal: Enhancing semi-supervision via
  label grouping and co-training.
\newblock In \emph{Proceedings of the IEEE/CVF Conference on Computer Vision
  and Pattern Recognition}, pp.\  7241--7250, 2021.

\bibitem[Nassar et~al.(2023)Nassar, Hayat, Abbasnejad, Rezatofighi, and
  Haffari]{nassar2023protocon}
Nassar, I., Hayat, M., Abbasnejad, E., Rezatofighi, H., and Haffari, G.
\newblock Protocon: Pseudo-label refinement via online clustering and
  prototypical consistency for efficient semi-supervised learning.
\newblock In \emph{Proceedings of the IEEE/CVF Conference on Computer Vision
  and Pattern Recognition}, pp.\  11641--11650, 2023.

\bibitem[Pham et~al.(2021)Pham, Dai, Xie, and Le]{pham2021meta}
Pham, H., Dai, Z., Xie, Q., and Le, Q.~V.
\newblock Meta pseudo labels.
\newblock In \emph{Proceedings of the IEEE/CVF Conference on Computer Vision
  and Pattern Recognition}, pp.\  11557--11568, 2021.

\bibitem[Radford et~al.(2021)Radford, Kim, Hallacy, Ramesh, Goh, Agarwal,
  Sastry, Askell, Mishkin, Clark, et~al.]{radford2021learning}
Radford, A., Kim, J.~W., Hallacy, C., Ramesh, A., Goh, G., Agarwal, S., Sastry,
  G., Askell, A., Mishkin, P., Clark, J., et~al.
\newblock Learning transferable visual models from natural language
  supervision.
\newblock In \emph{International conference on machine learning}, pp.\
  8748--8763. PMLR, 2021.

\bibitem[Rasmus et~al.(2015{\natexlab{a}})Rasmus, Berglund, Honkala, Valpola,
  and Raiko]{rasmus2015semi}
Rasmus, A., Berglund, M., Honkala, M., Valpola, H., and Raiko, T.
\newblock Semi-supervised learning with ladder networks.
\newblock \emph{Advances in Neural Information Processing Systems},
  28:\penalty0 3546--3554, 2015{\natexlab{a}}.

\bibitem[Rasmus et~al.(2015{\natexlab{b}})Rasmus, Valpola, Honkala, Berglund,
  and Raiko]{Rasmus2015SemiSupervisedLW}
Rasmus, A., Valpola, H., Honkala, M., Berglund, M., and Raiko, T.
\newblock Semi-supervised learning with ladder network.
\newblock \emph{ArXiv}, abs/1507.02672, 2015{\natexlab{b}}.

\bibitem[Shi et~al.(2023)Shi, Zhang, Zhen, Fan, and Yan]{shi2023understanding}
Shi, L., Zhang, G., Zhen, H., Fan, J., and Yan, J.
\newblock Understanding and generalizing contrastive learning from the inverse
  optimal transport perspective.
\newblock 2023.

\bibitem[Sohn et~al.(2020)Sohn, Berthelot, Carlini, Zhang, Zhang, Raffel,
  Cubuk, Kurakin, and Li]{sohn2020fixmatch}
Sohn, K., Berthelot, D., Carlini, N., Zhang, Z., Zhang, H., Raffel, C.~A.,
  Cubuk, E.~D., Kurakin, A., and Li, C.-L.
\newblock Fixmatch: Simplifying semi-supervised learning with consistency and
  confidence.
\newblock \emph{Advances in neural information processing systems},
  33:\penalty0 596--608, 2020.

\bibitem[Taherkhani et~al.(2020)Taherkhani, Dabouei, Soleymani, Dawson, and
  Nasrabadi]{taherkhani2020transporting}
Taherkhani, F., Dabouei, A., Soleymani, S., Dawson, J., and Nasrabadi, N.~M.
\newblock Transporting labels via hierarchical optimal transport for
  semi-supervised learning.
\newblock In \emph{European Conference on Computer Vision}, pp.\  509--526.
  Springer, 2020.

\bibitem[Tai et~al.(2021)Tai, Bailis, and Valiant]{tai2021sinkhorn}
Tai, K.~S., Bailis, P.~D., and Valiant, G.
\newblock Sinkhorn label allocation: Semi-supervised classification via
  annealed self-training.
\newblock In \emph{International Conference on Machine Learning}, pp.\
  10065--10075. PMLR, 2021.

\bibitem[Tan et~al.(2023{\natexlab{a}})Tan, Wang, and Zhang]{tan2023seal}
Tan, Z., Wang, Z., and Zhang, Y.
\newblock Seal: Simultaneous label hierarchy exploration and learning.
\newblock \emph{arXiv preprint arXiv:2304.13374}, 2023{\natexlab{a}}.

\bibitem[Tan et~al.(2023{\natexlab{b}})Tan, Yang, Huang, Yuan, and
  Zhang]{tan2023information}
Tan, Z., Yang, J., Huang, W., Yuan, Y., and Zhang, Y.
\newblock Information flow in self-supervised learning.
\newblock \emph{arXiv preprint arXiv:2309.17281}, 2023{\natexlab{b}}.

\bibitem[Tan et~al.(2023{\natexlab{c}})Tan, Zhang, Yang, and
  Yuan]{tan2023contrastive}
Tan, Z., Zhang, Y., Yang, J., and Yuan, Y.
\newblock Contrastive learning is spectral clustering on similarity graph.
\newblock \emph{arXiv preprint arXiv:2303.15103}, 2023{\natexlab{c}}.

\bibitem[Tarvainen \& Valpola(2017)Tarvainen and Valpola]{tarvainen2017mean}
Tarvainen, A. and Valpola, H.
\newblock Mean teachers are better role models: Weight-averaged consistency
  targets improve semi-supervised deep learning results.
\newblock \emph{Advances in neural information processing systems}, 30, 2017.

\bibitem[Tian et~al.(2020)Tian, Krishnan, and Isola]{tian2020contrastive}
Tian, Y., Krishnan, D., and Isola, P.
\newblock Contrastive multiview coding.
\newblock In \emph{Computer Vision--ECCV 2020: 16th European Conference,
  Glasgow, UK, August 23--28, 2020, Proceedings, Part XI 16}, pp.\  776--794.
  Springer, 2020.

\bibitem[Tschannen et~al.(2019)Tschannen, Djolonga, Rubenstein, Gelly, and
  Lucic]{tschannen2019mutual}
Tschannen, M., Djolonga, J., Rubenstein, P.~K., Gelly, S., and Lucic, M.
\newblock On mutual information maximization for representation learning.
\newblock \emph{arXiv preprint arXiv:1907.13625}, 2019.

\bibitem[Vaswani et~al.(2017)Vaswani, Shazeer, Parmar, Uszkoreit, Jones, Gomez,
  Kaiser, and Polosukhin]{vaswani2017attention}
Vaswani, A., Shazeer, N., Parmar, N., Uszkoreit, J., Jones, L., Gomez, A.~N.,
  Kaiser, {\L}., and Polosukhin, I.
\newblock Attention is all you need.
\newblock \emph{Advances in neural information processing systems}, 30, 2017.

\bibitem[Wang et~al.(2022{\natexlab{a}})Wang, Lukasiewicz, Massiceti, Hu,
  Pavlovic, and Neophytou]{wang2022np}
Wang, J., Lukasiewicz, T., Massiceti, D., Hu, X., Pavlovic, V., and Neophytou,
  A.
\newblock Np-match: When neural processes meet semi-supervised learning.
\newblock In \emph{International Conference on Machine Learning}, pp.\
  22919--22934. PMLR, 2022{\natexlab{a}}.

\bibitem[Wang et~al.(2021)Wang, Lian, and Yu]{wang2021data}
Wang, X., Lian, L., and Yu, S.~X.
\newblock Data-centric semi-supervised learning.
\newblock \emph{arXiv preprint arXiv:2110.03006}, 2021.

\bibitem[Wang et~al.(2022{\natexlab{b}})Wang, Wu, Lian, and
  Yu]{wang2022debiased}
Wang, X., Wu, Z., Lian, L., and Yu, S.~X.
\newblock Debiased learning from naturally imbalanced pseudo-labels.
\newblock In \emph{Proceedings of the IEEE/CVF Conference on Computer Vision
  and Pattern Recognition}, pp.\  14647--14657, 2022{\natexlab{b}}.

\bibitem[Wang et~al.(2022{\natexlab{c}})Wang, Chen, Fan, SUN, Tao, Hou, Wang,
  Yang, Zhou, Guo, Qi, Wu, Li, Nakamura, Ye, Savvides, Raj, Shinozaki, Schiele,
  Wang, Xie, and Zhang]{usb}
Wang, Y., Chen, H., Fan, Y., SUN, W., Tao, R., Hou, W., Wang, R., Yang, L.,
  Zhou, Z., Guo, L.-Z., Qi, H., Wu, Z., Li, Y.-F., Nakamura, S., Ye, W.,
  Savvides, M., Raj, B., Shinozaki, T., Schiele, B., Wang, J., Xie, X., and
  Zhang, Y.
\newblock {USB}: A unified semi-supervised learning benchmark for
  classification.
\newblock In \emph{Thirty-sixth Conference on Neural Information Processing
  Systems Datasets and Benchmarks Track}, 2022{\natexlab{c}}.

\bibitem[Wang et~al.(2022{\natexlab{d}})Wang, Chen, Heng, Hou, Savvides,
  Shinozaki, Raj, Wu, and Wang]{wang2022freematch}
Wang, Y., Chen, H., Heng, Q., Hou, W., Savvides, M., Shinozaki, T., Raj, B.,
  Wu, Z., and Wang, J.
\newblock Freematch: Self-adaptive thresholding for semi-supervised learning.
\newblock \emph{arXiv preprint arXiv:2205.07246}, 2022{\natexlab{d}}.

\bibitem[Xie et~al.(2020)Xie, Dai, Hovy, Luong, and Le]{xie2020unsupervised}
Xie, Q., Dai, Z., Hovy, E., Luong, T., and Le, Q.
\newblock Unsupervised data augmentation for consistency training.
\newblock \emph{Advances in Neural Information Processing Systems},
  33:\penalty0 6256--6268, 2020.

\bibitem[Xu et~al.(2021)Xu, Shang, Ye, Qian, Li, Sun, Li, and Jin]{xu2021dash}
Xu, Y., Shang, L., Ye, J., Qian, Q., Li, Y.-F., Sun, B., Li, H., and Jin, R.
\newblock Dash: Semi-supervised learning with dynamic thresholding.
\newblock In \emph{International Conference on Machine Learning}, pp.\
  11525--11536. PMLR, 2021.

\bibitem[Zhang et~al.(2021)Zhang, Wang, Hou, Wu, Wang, Okumura, and
  Shinozaki]{zhang2021flexmatch}
Zhang, B., Wang, Y., Hou, W., Wu, H., Wang, J., Okumura, M., and Shinozaki, T.
\newblock Flexmatch: Boosting semi-supervised learning with curriculum pseudo
  labeling.
\newblock \emph{Advances in Neural Information Processing Systems},
  34:\penalty0 18408--18419, 2021.

\bibitem[Zhang et~al.(2023{\natexlab{a}})Zhang, Tan, Yang, Huang, and
  Yuan]{zhang2023matrix}
Zhang, Y., Tan, Z., Yang, J., Huang, W., and Yuan, Y.
\newblock Matrix information theory for self-supervised learning.
\newblock \emph{arXiv preprint arXiv:2305.17326}, 2023{\natexlab{a}}.

\bibitem[Zhang et~al.(2023{\natexlab{b}})Zhang, Yang, Tan, and
  Yuan]{zhang2023relationmatch}
Zhang, Y., Yang, J., Tan, Z., and Yuan, Y.
\newblock Relationmatch: Matching in-batch relationships for semi-supervised
  learning.
\newblock \emph{arXiv preprint arXiv:2305.10397}, 2023{\natexlab{b}}.

\bibitem[Zheng et~al.(2021)Zheng, You, Wang, Qian, Zhang, Wang, and
  Xu]{zheng2021ressl}
Zheng, M., You, S., Wang, F., Qian, C., Zhang, C., Wang, X., and Xu, C.
\newblock Ressl: Relational self-supervised learning with weak augmentation.
\newblock \emph{Advances in Neural Information Processing Systems},
  34:\penalty0 2543--2555, 2021.

\bibitem[Zheng et~al.(2022)Zheng, You, Huang, Wang, Qian, and
  Xu]{zheng2022simmatch}
Zheng, M., You, S., Huang, L., Wang, F., Qian, C., and Xu, C.
\newblock Simmatch: Semi-supervised learning with similarity matching.
\newblock In \emph{Proceedings of the IEEE/CVF Conference on Computer Vision
  and Pattern Recognition}, pp.\  14471--14481, 2022.

\end{thebibliography}
\bibliographystyle{icml2024}

\clearpage
\appendix
\onecolumn

\begin{center}
\textbf{\Large Appendix}
\end{center}

\section{More on proofs}

\subsection{Proof of lemma \ref{mean lemma}}

\begin{proof}
By using the definition of Wasserstein distance. We find that $\mathcal{W}(\delta_x, \sum^m_{i=1} \frac{1}{m} \delta_{s_i} ) = \frac{1}{m} \sum^{m}_{i=1} (x-s_i)^2$. As this is a quadratic function of $x$, we can immediately derive that the unique minimizer is $\frac{\sum^{m}_{i=1} s_i}{m}$.    
\end{proof}

\subsection{Proof of lemma \ref{W dis}}

\begin{proof}
Note $\mu(i) \leq \nu(i)$ for any $i \neq k$, Thus by the probability constraints we know that $\mu(k) \geq \nu(k)$. As the cost is generated by a metric, we know that $\mathbf{C}_{kk}=0$. Consider transporting mass from $\nu$ to $\mu$, as the cost from $k$ to $k$ is 0, $\nu$ will transport all $\nu(k)$ to $\mu(k)$. For any $i \neq k$, if $\nu$ transport $\Delta>0$ mass to point $j$ ($j \neq k$). Then as $\nu(j) \geq \mu(j)$, $\nu$ can only transport the mass $\Delta$ to the unique point where $\nu$ has smaller mass than $\mu$.  From triangular inequality $\mathbf{C}_{ik} \leq \mathbf{C}_{ij} + \mathbf{C}_{jk}$, this is costly than transporting directly from $i$ to $k$. Thus the optimal plan is to transport all the residual mass $\nu(i) - \mu(i)$ to node $k$. Thus the conclusion follows.  
\end{proof}

\section{More algorithms derived from semantic-distribution matching with  optimal transport} \label{More algorithms}
\citet{shi2023understanding} show that SimCLR \citep{chen2020simple} and MoCo \citep{he2019momentum} can be understood by the optimal transport viewpoint. We would like to show that many other import algorithms can also be derived from optimal transport. Other understandings of self-supervised learning can be found in \citep{tan2023contrastive, tan2023information, zhang2023matrix}.

\subsection{Cross-entropy based contrastive methods}

SimMatch \citep{zheng2022simmatch}, CoMatch \citep{li2021comatch}, ReSSL \citep{zheng2021ressl}, SwAV  \citep{caron2020unsupervised} and DINO \citep{caron2021emerging} adopt the teacher student setting and use KL divergence in their loss (consider the effect of stop-gradient, the cross-entropy is equivalent to KL divergence). Similar to the derivation in Section \ref{OT understand}, the teacher (student) matching matrices are generated by setting the cost matrix using the (negative) similarity of query samples between buffer samples (SimMatch, ReSSL), class prototypes (SwAV), other samples in a batch (CoMatch) or classification head weights (DINO). The derivation is similar to Section \ref{OT understand}.

While both our OTMatch and contrastive learning-based methods consider the relationship between classes, there are some crucial distinctions. Our OTMatch focuses on aligning the class classification probabilities of two augmented views, following the line of work such as FixMatch, FlexMatch, and FreeMatch. In contrast, contrastive learning-based methods emphasize the consistency between two batches of augmented views. To be more specific, our OTMatch calculates each optimal transport loss exclusively involving the two augmented views. In contrast, contrastive learning-based methods such as SimMatch and CoMatch align the two batches by utilizing the representation similarity between samples in the batch. As a result, contrastive learning-based methods necessitate an additional branch, apart from the one calculating sample-wise consistency. Therefore, our OTMatch is orthogonal to contrastive learning-based methods and can be combined with them.

\subsection{CLIP}

CLIP \citep{radford2021learning} is a multi-modal learning algorithm. For a batch of image text pairs $\{ (I_i, T_i) \}^B_{i=1}$, the image to text loss is as follows:
\begin{equation}
\mathcal{L}_{I \rightarrow T} =   - \sum^{B}_{i=1} \log \frac{\exp(\langle f_I(I_i), f_T(T_i) \rangle / \tau)}{\sum^{B}_{k=1} \exp(\langle f_I(I_i), f_T(T_k) \rangle / \tau)},
\end{equation}
where $f_I$ is the image encoder and $f_T$ is the text encoder and $\tau$ is the temperature.

The loss can be retrieved by setting the cost matrix as $\mathbf{C}_{i, j}= \| f_I(I_i) - f_T(T_j)\|^2_2$ and the ground truth matching matrix $\bar{\mathbf{T}}= \text{diag}{\frac{1}{B} \mathbf{1}_B}$. By noticing the fact that representations are normalized and using equation (\ref{matching matrix}), calculating the loss in IOT will give the loss $\mathcal{L}_{I \rightarrow T}$. Different from the uni-modal case where there will only be transportation between the single modality. In multi-modal cases, there will also exist a symmetric transportation loss $\mathcal{L}_{T \rightarrow I}$, which can be explained by optimal transport similarly.

\subsection{SupCon}

SupCon \citep{khosla2020supervised} is a supervised learning method that generates compact representations of images by incorporating label information. Suppose $I$ a batch of augmented images and $ A(i)= I-\{i\}$. Denote $\mathbf{z}_i$ as the representation of image $i$, its label is $\tilde{\boldsymbol{y}}_i$.

\begin{equation}
\mathcal{L}_{\text {SupCon }}=\sum_{i \in I}-\frac{1}{|P(i)|}\log  \sum_{p \in P(i)} \frac{\exp \left(\mathbf{z}_i \cdot \mathbf{z}_p / \tau\right)}{\sum_{a \in A(i)} \exp \left(\mathbf{z}_i \cdot \mathbf{z}_a / \tau\right)}.
\end{equation}

Here, $P(i) \equiv\left\{p \in A(i): \tilde{\boldsymbol{y}}_p=\tilde{\boldsymbol{y}}_i\right\}$ is the set of indices of all positives in the batch.

By setting $\mathbf{C}_{ii} = + \infty$ and $\mathbf{C}_{ij} = c - \mathbf{z}_i \cdot \mathbf{z}_j $. Noticing that the $i$-th row of ground truth matching matrix $\bar{\mathbf{T}}_{ij}= \frac{1}{I |P(i)|} \delta^{\tilde{\boldsymbol{y}}_j}_{\tilde{\boldsymbol{y}}_i}$. By noticing the fact that representations are normalized and using equation (\ref{matching matrix}), calculating the loss in IOT will give the SupCon loss.

\subsection{BYOL and SimSiam}

BYOL \citep{grill2020bootstrap} and SimSiam \citep{chen2021exploring} uses the MSE loss between two augmented views. For a batch of images $\{ \mathbf{x}_i \}^B_{i=1}$, we usually apply different augmentations to the images and get two batches of representations $\{ \mathbf{z}^{(1)}_i \}^B_{i=1}$ and $\{ \mathbf{z}^{(2)}_i \}^B_{i=1}$.

Take $\mu = \frac{1}{B} \mathbf{1}_B$, \citet{shi2023understanding} using the optimal value of the following optimization problem \ref{IOT 1} to explain SimCLR and MoCo. We consider first change the inner minimization of the entropic regularization problem in \ref{IOT 1} into the common optimal transport problem and get optimization problem  \ref{IOT 2}. However, this bi-level optimization problem is still hard to solve. Thus we then relax the problem into an optimization problem \ref{IOT 3}.

Take the ground-truth matching matrix as $\bar{\mathbf{T}} = \text{diag}{\frac{1}{B} \mathbf{1}_B}$. The cost matrix $\mathbf{C}_{i, i}= \| \mathbf{z}^{(1)}_i - \mathbf{z}^{(2)}_i \|^2_2$ and $\mathbf{C}_{i, j} = c +  \| \mathbf{z}^{(1)}_i - \mathbf{z}^{(2)}_i \|^2_2$ ($j \neq i$), where $c$ is a large constant.

Then the optimization problem will be $\text{Const.} + \frac{1}{B} \sum_{i} - \log \mathbf{T}_{ii} + \lambda \sum_{i} (\mathbf{C}_{i, i} \mathbf{T}_{i, i} + \sum_{j \neq i} (c+\mathbf{C}_{i, i}) \mathbf{T}_{i, j})$. Using the constraint of $U(\mu)$ and simplifying the constant out, the objective function will be $\frac{1}{B} \sum_i (-Bc\lambda \mathbf{T}_{i, i} - \log \mathbf{T}_{i, i}) + \lambda \sum_{i} \mathbf{C}_{i, i} +\frac{c\lambda}{B} $. As $\mathbf{T} \in U(\mu)$, the optimal value is $\text{Const.} + \lambda \sum_i \mathbf{C}_{ii}$. This exactly recovers the MSE loss.

\begin{align} \label{IOT 1}
\min \quad & \text{KL}(\bar{\mathbf{T}} \| \mathbf{T}^{\theta}) \nonumber\\
\text{subject to}\quad & \mathbf{T}^{\theta} = \argmin_{\mathbf{T} \in U(\mu)} \langle \mathbf{C}^{\theta}, \mathbf{T} \rangle - \epsilon \text{H}(\mathbf{T}). 
\end{align}

\begin{align} \label{IOT 2}
\min \quad & \text{KL}(\bar{\mathbf{T}} \| \mathbf{T}^{\theta}) \nonumber\\
\text{subject to}\quad & \mathbf{T}^{\theta} = \argmin_{\mathbf{T} \in U(\mu)} \langle \mathbf{C}^{\theta}, \mathbf{T} \rangle . 
\end{align}

\begin{align} \label{IOT 3}
\min \quad & \text{KL}(\bar{\mathbf{T}} \| \mathbf{T}) + \lambda \langle \mathbf{C}, \mathbf{T} \rangle  \nonumber\\
\text{subject to}\quad & \mathbf{T} \in U(\mu). 
\end{align}

\subsection{Integrating OTMatch with self-supervised learning method}

We also provide a solution for applying our training algorithm to self-supervised learning methods like DINO~\citep{dino}.
Given probability distributions $p$ (teacher distribution) and $q$ (student distribution), the cross-entropy (CE) loss is defined as $- \sum p_i \log q_i$.
As $p$ uses sharpening, the proxy losses can be defined in the following way:

\begin{equation*}
\sum^{K}_{i=1} \mathbf{C}_{ik} \mid p_i - q_i \mid, \text{where } k = \argmax p.      
\end{equation*}

where $\mathbf{C}$ represents the cost in optimal transport. 
The results are shown in Table \ref{dino}.

\begin{table}[t]
\renewcommand{\arraystretch}{0.8}
\setlength{\tabcolsep}{6pt}
\centering
\caption{Accuracy ($\%$) on CIFAR-10.}
\label{dino}
\small
\begin{tabular}{cc}
\toprule
DINO   & 88.06   \\ \midrule
DINO+OTMatch & \textbf{88.20} \\ \bottomrule
\end{tabular}
\end{table}

\section{Analysis of the running time}

\begin{table}[t]
\renewcommand{\arraystretch}{0.8}
\setlength{\tabcolsep}{6pt}
\centering
\caption{Per-iteration running time}
\label{running time}
\small
\begin{tabular}{cc}
\toprule
FreeMatch   & 0.23 s   \\ \midrule
OTMatch & 0.25 s \\ \bottomrule
\end{tabular}
\end{table}
We calculate the per-iteration running time of FreeMatch and OTMatch on CIFAR-10 with 40 labels. From Table \ref{running time}, it can be observed that our method introduces a small computation overhead.

\end{document}